\documentclass[sigconf]{acmart}
\usepackage{color}
\usepackage{amsmath}
\usepackage{multirow}
\usepackage{booktabs}
 \usepackage{array}

\AtBeginDocument{%
  \providecommand\BibTeX{{%
    \normalfont B\kern-0.5em{\scshape i\kern-0.25em b}\kern-0.8em\TeX}}}

\begin{document}

\author{Yicheng Guo}
\email{guoyicheng3@huawei.com}
\affiliation{%
  \institution{Huawei Technologies Co., Ltd}}
 
\author{Yujin Wen}
\email{wenyujin@huawei.com}
\affiliation{%
  \institution{Huawei Technologies Co., Ltd}} 

\author{Congwei Jian}
\email{jiangcongwei1@huawei.com}
\affiliation{%
  \institution{Huawei Technologies Co., Ltd}} 
  
\author{Yixin Lian}
\email{lianyixin1@huawei.com}
\affiliation{%
  \institution{Huawei Technologies Co., Ltd}}

\author{Yi Wan\textsuperscript{*}}
\email{wanyi20@huawei.com}
\affiliation{%
  \institution{Huawei Technologies Co., Ltd}}

\title{Detecting Log Anomalies with Multi-Head Attention (LAMA)}

\begin{abstract}
  Anomaly detection is a crucial and challenging subject that has been studied within diverse research areas. In this work, we explore the task of log anomaly detection (especially computer system logs and user behavior logs) by analyzing logs' sequential information. We propose LAMA, a multi-head attention based sequential model to process log streams as template activity (event) sequences. 
  A next event prediction task is applied to train the model for anomaly detection. Extensive empirical studies demonstrate that our new model outperforms existing log anomaly detection methods including statistical and deep learning methodologies, which validate the effectiveness of our proposed method in learning sequence patterns of log data.
\end{abstract}


\maketitle

\section{Introduction}
 Anomaly detection is a crucial and challenging subject that has been studied within diverse research areas, e.g., cyber security, finance, recommendation and advertising. Log data, especially computer system logs and user behavior logs, records significant events that are universally available everywhere and are valuable resources for understanding system status and users' intention. Logs are therefore one of the most essential sources for anomaly detection to build a secure and trustworthy environment. 

Generally, log anomaly detection involves three steps: log parsing, session creation, and anomaly detection. The purpose of log parsing is transferring unstructured log data into structured forms and extracting log templates (events) \cite{2016experience}. The state-of-the -art log parsing method is represented by Spell \cite{2019spell}. 
As for session creation, log data are usually grouped by one of the three following windows: fixed window, sliding window, and session window. 
In this study, we choose session window since the HDFS dataset used in this paper is based on identifiers (block\_id), recording system operations such as allocation, writing, replication. Log sessions are then  feed into anomaly detection models.

Existing approaches for log anomaly detection can be classified into two groups. The first group of approaches use quantitative information of log events. Program execution and user behaviors have quantitative relationships. If such relationships are broken, we say there are anomalies in logs. \cite{2008if} proposed a tree-based method and \cite{2009PCA} proposed a Principal Component Analysis (PCA)-based method; by mining invariant patterns between log events, \cite{2010im} proposed invariant mining model; \cite{2016lg} used hierarchical clustering to detect small clusters as anomalies. The second group of approaches are based on sequential information. These approaches assume programs execution or user behaviors follow fixed flows or have intrinsic patterns. If a log sequence deviates from normal order of flows, it is treated as sequential anomaly. For example, \cite{spectrogram} proposed a mixture-of-Markov-Chains model by learning a proper distribution on overlapping $n$-grams; \cite{2016zhang} and \cite{2017deeplog} utilize LSTM \cite{lstm}, a deep learning model, to predict next log event. In this paper, we focus on the detection of sequential anomalies.

However, above models have some drawbacks. For example, 
statistical models ignore order of log sequences; Markov Chain based models assume that every event in logs only influenced by its recent $N$ predecessors rather than the entire history and with the increase of $N$, models' computational complexity will grow exponentially; RNN based models  fail to explicitly capture fine-grained relationships between events in long sequences.

\begin{figure}[htbp]
  \centering
  \includegraphics[width=2.5in]{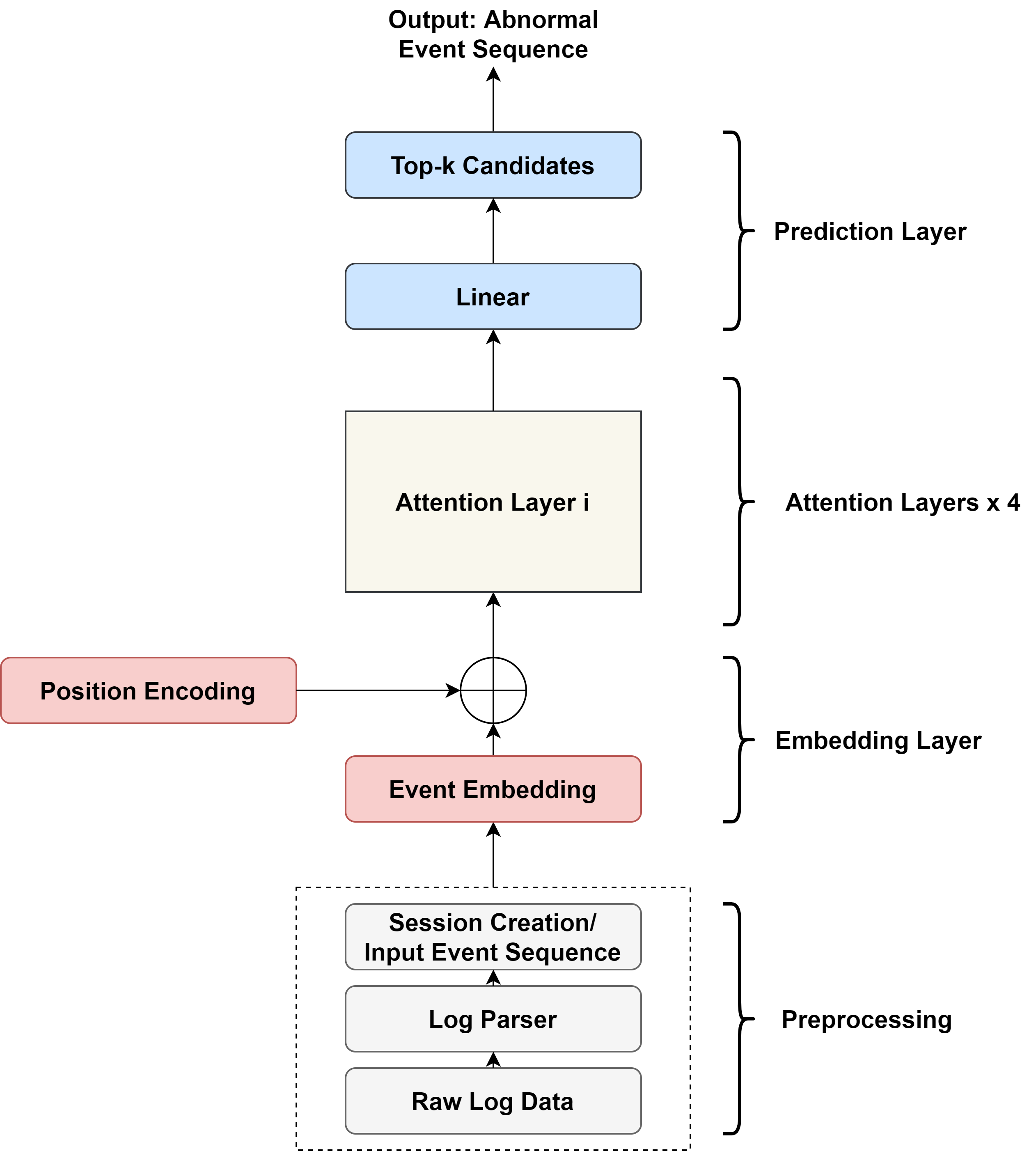}
  \caption{LAMA Architecture.}
  \Description{LAMA Architecture.}
  \label{F1}
\end{figure}

To address above drawbacks, we propose LAMA, a multi-head self-attention based sequential model to process log streams with sequential information. Attention mechanism learns to pay attention to the most important parts of the target. Standard vanilla attention often acts as an additional component to the original model (e.g. RNNs with attention, CNNs with attention). Our work introduces the multi-head attention mechanism proposed by Transfomer \cite{2017attention}, which is proved to be powerful in capturing sequences features in natural language scenarios.
The prime contributions of this work are as follows.
\begin{itemize}
    \item We proposed a novel self-attention based model for log anomaly detection. Different from previous works use vanilla attention as add-ons(e.g. \cite{10.1145/3217871.3217872}), LAMA uses attention to capture patterns in sequence data, without RNN based structures. 
    \item Compared with works applying seq2seq models to anomaly detection task such as \cite{8904299}, our model has a lightweight structure designed for the task. LAMA gets rid of the massive decoder, which makes the model more efficient.
\end{itemize}

The remainder of this paper is organized as follows: the methodology
is elaborated in Section 2, and then experimental results are presented in Section 3. Finally, we conclude our work in Section 4.

\section{Methodology}
In this section, we present the proposed multi-head self-attention based sequential model, named LAMA. The pipeline of LAMA is illustrated in Figure \ref{F1}. Preprocessing part is beyond the scope of this paper, so we focus on embedding layer, attention layer and prediction layer.
\subsection{Training Task Design}
\label{sec:task}
Suppose we obtain $n$ distinct events (templates) from log parsing, and $m$ sessions from session extraction. Let $E = \{e_1, e_2, \dots, e_n\}$ be the whole set of distinct events and $S_u = (s_1, s_2, \dots, s_{n_u})$ $(u = 1, 2, \dots, m)$ be the $u$\textsuperscript{th} session with $n_u$ events recorded, where $s_j \in E$ for $j = 1, 2, \dots, n_u$. Each session is converted to several fixed length sequences via sliding window (length $l$) method. For session $S_u$ with $n_u > l$, $n_u - l$ sequences are created, from $(s_1, s_2, \dots, s_l)$ to $(s_{n_u-l}, s_{n_u-l+1},$ $\dots, s_{n_u-1})$. Sessions with $n_u \le l$ are padded to length of $l$.

The model takes each sequence as input to predict the event next to it. If the truth event is not included in top-$k$ predicted candidates, the sequence is labelled as abnormal. A session is regarded as anomaly if it has abnormal sequences. 

 HDFS dataset contains several out of vocabulary (OOV) events, which are events never been seen during training. Many of previous deep-learning based studies use the whole set of events in prediction layer. However, one can never predict the number of OOV events in a real task. Besides, such a prediction gives different probabilities for different OOV events, which doesn't make sense. Our model predicts events appeared in training data only. Sessions with OOV events are labelled as anomaly directly.

\subsection{Our proposed model: LAMA}
\begin{figure}[ht]
  \centering
  \includegraphics[width=1.3in]{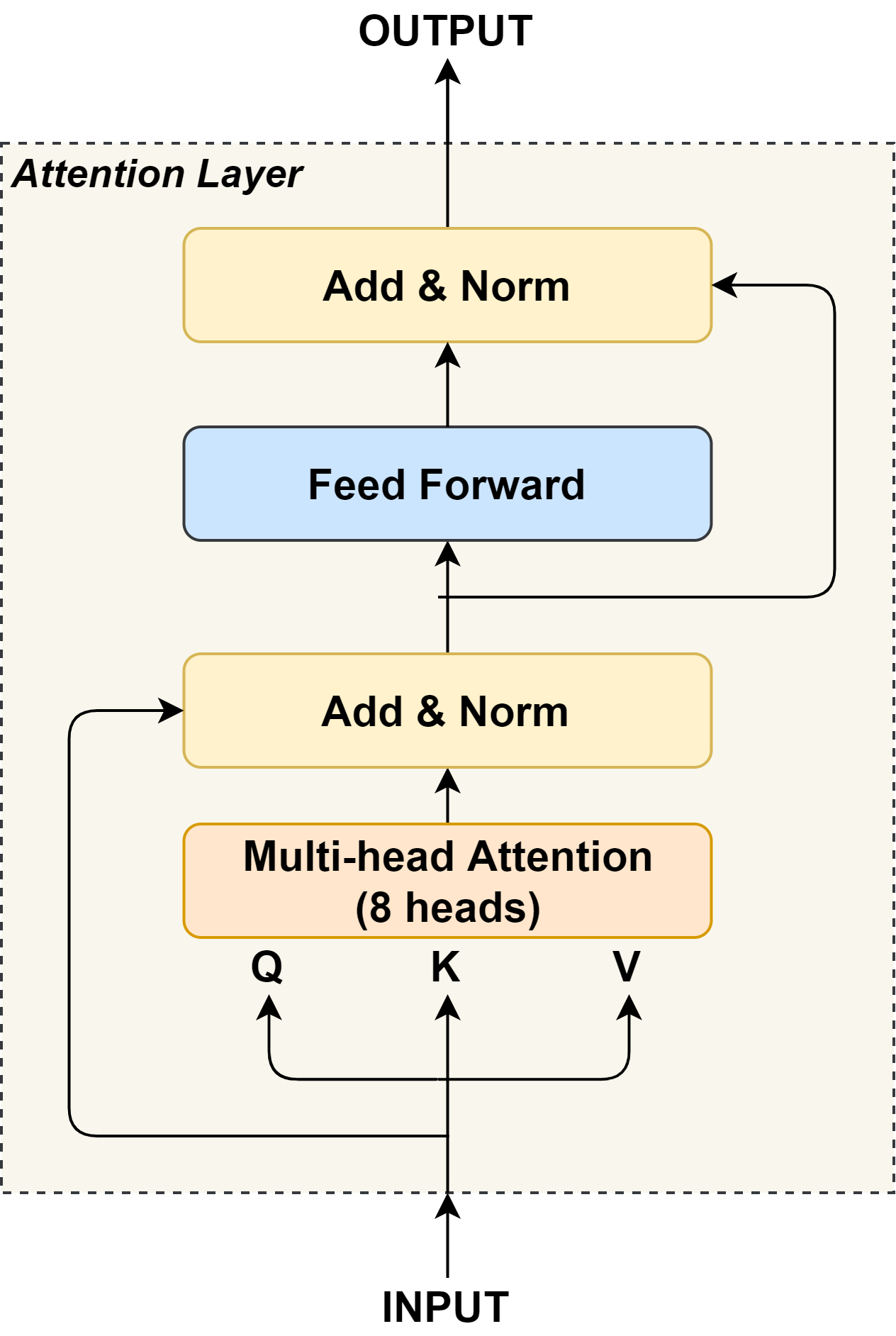}
  \caption{Attention Layer.}
  \Description{Attention Layer.}
  \label{F2}
\end{figure}
\textbf{Embedding Layer:} The embedding layer converts input events to vectors of dimension $d$ according to their ids. To make use of the order of sequences, this layer creates positional embedding vectors for events in each sequence based on their indices. The sum of these two embedding are considered as the embedding of a input sequence. Following \cite{2017attention}, we choose sine and cosine functions of different frequencies as the positional embedding functions. The embedding matrix for a input sequence is denoted as $X \in \mathbb{R}^{l \times d}$.\\

\textbf{Attention Layer:} The second component of LAMA is attention layer. In this layer, we choose the scaled dot-product attention introduced in \cite{2017attention}:
\begin{equation*}
    \text{Attention}(Q, K, V) = \text{softmax}\left(\frac{QK^{\top}}{\sqrt{d/h}}\right)V
\end{equation*}
where $h$ is the number of heads, and $Q$, $K$, $V$ represent queries, keys and values, respectively. This study adopts self-attention mechanism with $Q = K = V = X$. With multi-heads attention and point-wise feed-forward layer, LAMA builds deep representations for event sequences,
\begin{gather*}
    S = \text{MultiHead}(X) = \text{Concat}(\text{head}_1, \dots, \text{head}_h)W^{O},\\
    \text{head}_i = \text{Attention}(XW^{Q}_i, XW^{K}_i, XW^{V}_i),~i = 1, \dots, h,\\
    \text{FFN}(S) = \text{max}(0, SW_1 + b_1)W_2 + b_2,
\end{gather*}
where $W_1$, $W_2$, $W^{O} \in \mathbb{R}^{d \times d}$, $W^{Q}$, $W^{K}$, $W^{V} \in \mathbb{R}^{d \times d/h}$ and $b_1$, $b_2 \in \mathbb{R}^{d}$.
We stack $b$ self-Attention layers to learn complex event transition patterns.\\ 
\textbf{Prediction Layer:} The prediction layer is composed of a fully connected layer and softmax normalization. It makes prediction of next event for each sequence based on the sequence representation $\hat{X}$ provided by the attention layer:
\begin{equation*}
    p = \text{softmax}(\hat{X}W + b),
\end{equation*}
where $W$, $b$ denote $d \times d$, $d$-dimensional matrices in the linear layer.\\
\textbf{Model Learning: } 
Our model takes sequences as input, the predicted next event id as output, and we adopt cross-entropy loss as objective function:
\begin{equation*}
    -\sum_{u=1}^{m}\sum_{j=1}^{n_u-l}\sum_{k=1}^{n}\left[y_{u,j,k}\log(p_{u,j,k})\right],
\end{equation*}
where $y_{u, j, k}$ denotes the $k$\textsuperscript{th} candidate event for the $j$\textsuperscript{th} sequence in the $u$\textsuperscript{th} session, and $p_{u, j, k}$ is the corresponding predicted probability.

\section{Experiments}
This section describes the dataset and evaluation metrics employed for our experiments, followed by introduction of baseline models and implementation details. Performances of LAMA and an analysis of the model structure is then presented.
\subsection{Settings}
\textbf{Dataset Description: } 
We choose HDFS dataset for this study. HDFS dataset was generated and collected from Amazon EC2 platform through running Hadoop-based map-reduce jobs. It contains messages about blocks when they are allocated, replicated, written, or deleted. Each block is assigned with a unique ID in the log. Thus, we extract log sessions by block IDs. HDFS dataset contains $11,197,954$ log entries. After session extraction, we obtain $575,061$ sessions with $16,838$ marked as anomalies. \\
\textbf{Baselines: }
For comparison, we choose five statistical models and a deep learning model as baseline models to evaluate the performance of our proposed model, including Isolation Forest~\cite{2008if}, PCA~\cite{2009PCA}, Invariant Mining~\cite{2010im}, Log Clustering~\cite{2016lg}, $N$-gram~\cite{spectrogram}, and DeepLog~\cite{2017deeplog}.\\
\textbf{Data preprocessing: } Note that some of the statistical models are unsupervised that do not require training data,
whereas $N$-gram, DeepLog, and LAMA need a training set composed of normal records. In the following experiments, we leverage two different splitting ways, which take (1) $80\%$ of normal sessions as training data and the rest $20\%$ normal sessions along with all abnormal sessions as test data (2) $1\%$ of normal sessions as training data and the rest $99\%$ of normal data along with all abnormal sessions as test data. In the 80/20 splitting, As for the 1/99 splitting, we use data provided by DeepLog~\cite{2017deeplog}.\\

\textbf{Evaluation Metrics: }
In our experiments, we use Precision ($\frac{\text{TP}}{\text{TP}+\text{FP}}$), Recall ($\frac{\text{TP}}{\text{TP}+\text{FN}}$) and F$_1$-measure ($\frac{2\times\text{Precision}\times\text{Recall}}{\text{Precision}+\text{Recall}}$) to evaluate the performance of each model. Precision shows the percentage of how many reported anomalies are correct; Recall measures the percentage of how many anomalies are detected among true anomaly set; and F$_1$-measure is the harmonic mean of Precision and Recall.\\
\textbf{Implementation Details: }
The experiments are conducted on a Linux server with Intel(R) Xeon(R) 2.40GHz CPU and Tesla V100 GPU, with Python $3.6$ and pytorch $1.4.0$. Table~\ref{t3.2} summaries parameters for our model. 
\begin{table}[htbp!]
\centering
\caption{Parameters of LAMA}
\begin{tabular}{c|c|c|c}
\toprule
\multicolumn{4}{c}{Parameters of LAMA}\\
\midrule
head number ($h$) & 2/4/\textbf{8} & epochs & 5\\
self-attention layer ($b$) & 2/\textbf{4}/8 & batch size& 128 \\
embedding size ($d$) & 256 & dropout & 0.1\\
feed-forward size & 1024 & \# of templates (events) & 29 \\
top $k$ candidates & 4/6 & sequence length ($l$) & 10 \\
\bottomrule
\end{tabular}
\label{t3.2}
\end{table}

\subsection{Performance}
Table \ref{t3.3} presents the performance of LAMA and baseline models. Among these models, LAMA achieves the best score with F$_1$- measure of $0.985$ and $0.946$ for the two data splitting methods, respectively. For Deeplog and LAMA, the two numbers for each metric are results without and with OOV events, as described in Section~\ref{sec:task}. The first number without OOV events gives a better estimation of the performance of the model, since OOV events are not predictable. The second number with OOV events is used for comparison with other models. In 1/99 splitting, since test set contains a lot of OOV events, scores change a lot. In 80/20 splitting, however, the OOV events has little impact. 
The results reveals that, the multi-head attention is powerful to find out intrinsic patterns of input sequences, and relations between log events. Besides, we find LAMA converges faster than LSTM base model. In the 80/20 splitting, with 1 Tesla-V100 GPU, it takes less than one hour to train LAMA. We believe such a mechanism can be widely applied to analyze sequential structured data instead of LSTM.

\begin{table}[htbp!]
    \caption{Evaluation on HDFS logs.}
    \centering
    \begin{tabular}{p{1.3cm}<{\centering}p{1.2cm}p{1.3cm}p{1.3cm}p{1.4cm}}
    \toprule
    Model&Dataset Splitting &Precision&Recall&F$_1$-measure\\
    \midrule
    \multirow{2}*{\centering \shortstack{Isolation\\Forest}} 
     &$80/20$ &0.950 & 0.756 & 0.842 \\
     &$1/99$ & 0.799 & 0.613 & 0.694 \\
    \midrule
    \multirow{2}*{\centering PCA} 
     &$80/20$&0.974 & 0.445 & 0.611 \\
     &$1/99$&0.966 & 0.473 & 0.636 \\
    \midrule
    \multirow{2}*{\shortstack{Invariant\\Mining}} 
     &$80/20$&0.975 & 0.633  & 0.768  \\
     &$1/99$&0.893 &1.000 & 0.944\\
    \midrule
    \multirow{2}*{\shortstack{Log\\Clustering}} 
    &$80/20$&0.498 &0.530 &0.513  \\
     &$1/99$&0.330 &0.525 &0.405 \\
    \midrule
    \multirow{2}*{\shortstack{$N$-gram}} 
    &$80/20$&0.674 &0.673 &0.673\\
    &$1/99$&0.418 &0.417 &0.417\\
    \midrule
    \multirow{2}*{\shortstack{DeepLog}}
    &$80/20$&0.971/0.972&0.977/0.978 &0.974/0.975\\
    &$1/99$&0.732/0.830 &0.963/0.980 &0.832/0.900\\
    \midrule
    \multirow{2}*\shortstack{LAMA}
    &$80/20$ & 0.976/0.977 & 0.995/0.995 & \textbf{0.985/0.986}  \\
    &$1/99$&0.872/0.917 &0.954/0.976 &\textbf{0.911/0.946}\\
    \bottomrule
    \end{tabular}
    \label{t3.3}
\end{table}

\subsection{Model Analysis}
The following paragraph present an in-depth model analysis, aiming to further understand the effectiveness of the model. Table~\ref{t3.5} shows the performance of LAMA with varying number of attention layers and heads, with 80/20 splitting. Numbers presented are results without OOV events since they have little impact in this scenario. For each setting, we performed ten experiments.\\
With $h = 8$, a $4$-layer model has similar performance to a $6$-layer one. When $b$ reduced to $2$, the performance dropped. This is because log event sequences are not as complex as natural language, so that $6$ layers of attention are redundant. Nevertheless, a shallow model ($2$-layers) lacks for the ability to extract intrinsic patterns of log sequence precisely. 
With a smaller numbers of attention heads, the model usually keeps good performance. However, scores drop dramatically in some experiments, which leads to big variances. LAMA with less heads are not as stable as an 8-heads one.

\begin{table}[htbp!]
\renewcommand\arraystretch{1.5}
\centering
\caption{Impact of numbers of attention layers (b) and heads (h) in LAMA}
\label{t3.5}
\begin{tabular}{ccccc}
\toprule
$b$ & $h$ & Precision & Recall & F$_1$-measure\\
\midrule
$2$ & $8$ &  $0.979^{+0.003}_{-0.001}$ & $0.976^{+0.012}_{-0.012}$ & $0.977^{+0.005}_{-0.005}$  \\
$4$ & $2$ & $0.975^{+0.001}_{-0.001}$ & $0.982^{+0.015}_{-0.049}$ & $0.978^{+0.008}_{-0.026}$ \\
$4$ & $4$ & $0.975^{+0.001}_{-0.003}$ & $0.978^{+0.018}_{-0.158}$ & $0.976^{+0.010}_{-0.087}$ \\
$4$ & 8 & $0.976^{+0.001}_{-0.002}$ & $0.995^{+0.003}_{-0.005}$ &  $0.985^{+0.001}_{-0.003}$ \\
$6$ & 8 & $0.974^{+0.002}_{-0.005}$ & $0.997^{+0.003}_{-0.010}$ & $0.985^{+0.002}_{-0.006}$ \\

\bottomrule
\end{tabular}
\end{table}

\section{Conclusion}
In this paper, we proposed LAMA, a novel self-attention based sequential model for log anomaly detection. LAMA uses the multi-head attention mechanism from transformer \cite{2017attention}, with a lightweight structure designed for log anomaly detection scenario. We demonstrate that the multi-head attention mechanism is efficient in extracting patterns from sequential log data. We show the superiority and effectiveness of LAMA by extensive experiments. Our next step is to improve the model by incorporating rich context information such as time stamps, action types, and device information.

\bibliographystyle{ACM-Reference-Format}
\bibliography{LAMA}


\end{document}